\pdfoutput=1

\documentclass[11pt]{article}

\usepackage{ACL2023}

\usepackage{times}
\usepackage{latexsym}

\usepackage[T1]{fontenc}

\usepackage[utf8]{inputenc}

\usepackage{microtype}

\usepackage{inconsolata}

\usepackage{amsmath}
\usepackage{graphicx}
\usepackage{float}
\usepackage{subcaption}
\usepackage{amsfonts}
\usepackage{multicol}
\usepackage[bottom]{footmisc}
\usepackage{color, colortbl}
\definecolor{Gray}{gray}{0.9}
\usepackage{arydshln}
\DeclareMathOperator*{\argmax}{arg\,max}

\usepackage{pifont}
\newcommand{\xmark}{\ding{55}}%

\usepackage{multirow, booktabs}
\usepackage{caption}

\title{\textsc{MQAG}: Multiple-choice Question Answering and Generation for\\Assessing Information Consistency in Summarization}

\author{Potsawee Manakul, Adian Liusie, Mark J. F. Gales \\
  ALTA Institute, Department of Engineering, University of Cambridge \\
  \texttt{pm574@cam.ac.uk, al826@cam.ac.uk, mjfg@eng.cam.ac.uk}}

\begin{document}
\maketitle
\begin{abstract}
State-of-the-art summarization systems can generate highly fluent summaries. These summaries, however, may contain factual inconsistencies and/or information not present in the source. Hence, an important component of assessing the quality of summaries is to determine whether there is information consistency between the source and the summary. Existing approaches are typically based on lexical matching or representation-based methods. In this work, we introduce an alternative scheme based on standard information-theoretic measures in which the information present in the source and summary is directly compared. We propose a Multiple-choice Question Answering and Generation framework, MQAG, which approximates the information consistency by computing the expected statistical distance between summary and source answer distributions over automatically generated multiple-choice questions. This approach exploits multiple-choice answer probabilities, as predicted answer distributions can be compared. We conduct experiments on four summary evaluation datasets: QAG-CNNDM/XSum, XSum-Hallucination, Podcast Assessment, and SummEval. Experiments show that MQAG, using models trained on SQuAD or RACE, outperforms existing evaluation methods on the majority of tasks.\footnote{Code and model weights are available at \url{https://github.com/potsawee/mqag0}.}
\end{abstract}

\section{Introduction}
\begin{figure}[t]
    \centering
\includegraphics[width=0.95\linewidth,keepaspectratio]{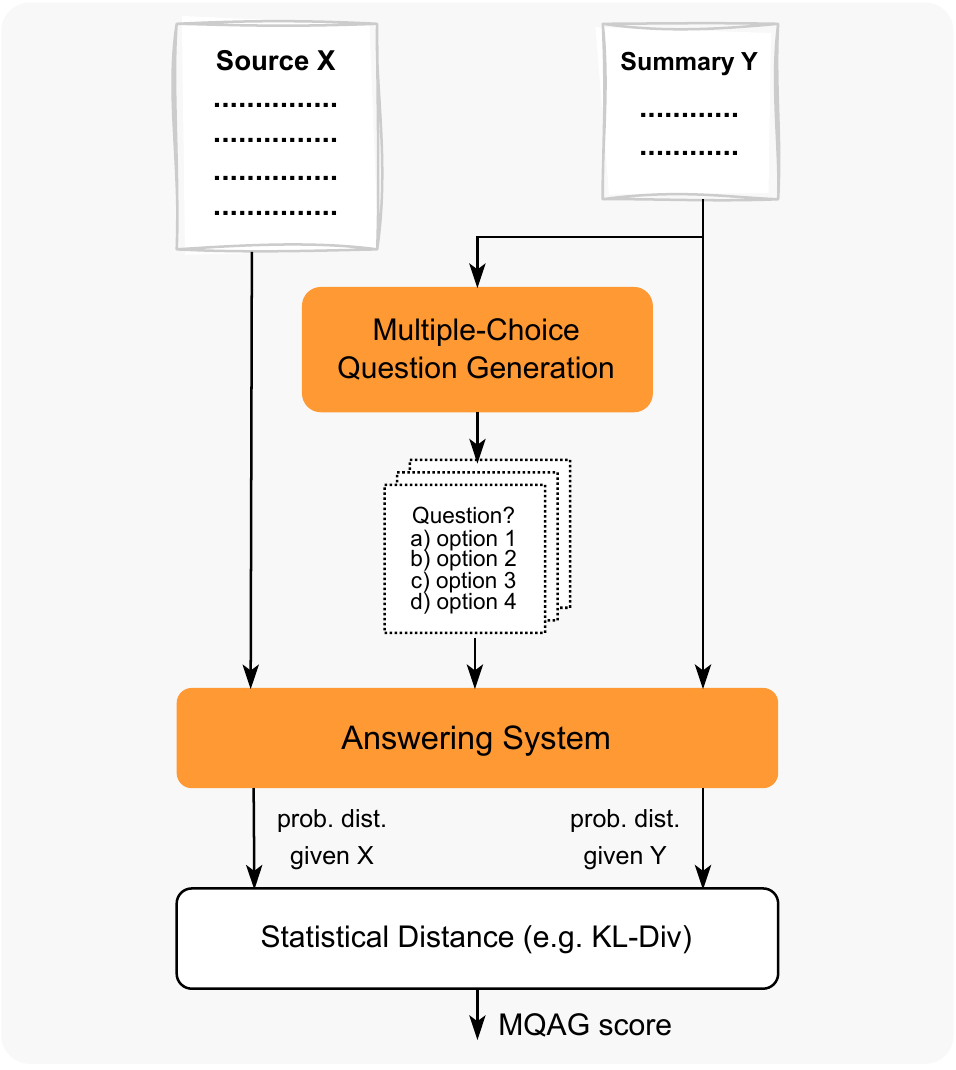}
    \caption{Multiple-choice Question Answering and Generation (MQAG) framework. The answers are represented by probability distributions over choices instead of text spans in existing question-answering approaches.}
    \label{fig:pipeline}    
\end{figure}

The objective of summary evaluation is to quantify the quality of summaries, either on a relative or an absolute scale. Accurate and reliable automatic summary evaluation systems are useful to researchers, as they provide an easy and cheap way to compare new summarization models to existing ones. Although current summarization systems have improved dramatically in the last decade, and are capable of generating highly fluent outputs \cite{lewis-etal-2020-bart, zhang2020pegasus, gpt3brown}, it has been shown that generated summaries are prone to exhibit factual errors or hallucinations \cite{kryscinski-etal-2019-neural, Huang2021TheFI, cao-etal-2022-hallucinated, Ji2022SurveyOH}. Thus, information consistency between the summary and source is an important assessment criterion.

Existing methods that measure information consistency generally perform lexical matching, either directly such as ROUGE \cite{lin-2004-rouge} and BLEU \cite{papineni-etal-2002-bleu}, or indirectly using more complex representations such as triple matching \cite{triple_matching}. Some recent approaches adopt question answering (QA) pipelines to detect factual inconsistencies \cite{chen2018semantic, wang-etal-2020-asking, durmus-etal-2020-feqa, deutsch-etal-2021-towards, nan-etal-2021-improving}. They are based on the assumption that if the source extracted answer is consistent with the summary extracted answer then the summary and source are consistent. The answers are compared using either lexical matching \cite{scialom-etal-2019-answers, wang-etal-2020-asking, durmus-etal-2020-feqa, scialom-etal-2021-questeval} or representation-based matching \cite{deutsch-roth-2022-benchmarking}. These span-based QA approaches may have lexical biases, and struggle with highly abstractive summaries or when dealing with multiple answer spans.

In this work, a measure of consistency between the source and summary is defined from an information-theoretic perspective. We propose a Multiple-choice Question Answering and Generation framework,  MQAG, where instead of comparing text-based answer spans, multiple-choice questions are generated and the resulting answer distributions from the source and summary are compared. The main contributions of this paper are: 
\begin{itemize}
    \item We provide an alternative and novel question answering-based approach for assessing information consistency. Our approach can represent the answers via probability distributions instead of lexical or embeddings.
    \item We show that our approach, MQAG, achieves state-of-the-art performance on four out of six summary evaluation tasks.
\end{itemize}

\section{Background and Related Work}
\label{section:existing_method}
Standard summary evaluation metrics such as ROUGE \cite{lin-2004-rouge} and METEOR \cite{banerjee-lavie-2005-meteor} are designed to assess summaries against ground-truth summaries, i.e. reference summaries. However, these metrics have been shown to have a low correlation with human judgements \cite{fabbri-etal-2021-summeval}. In practice, there is no ground-truth summary to be used as the reference, and evaluation methods need to compare the summary against the source. Therefore, the scope of this work is assessing the summary against the source. 

Although there are several aspects of good summaries, including fluency, coherency, coverage or consistency, generation systems are becoming much more capable of generating fluent texts, so the fluency/coherency aspects are less of a concern compared to consistency and hallucination problems \cite{ji2023hallucination}. Thus, this work focuses on \textit{consistency}. Because the definition of consistent information can depend on one's interpretation, we follow the definition of `faithfulness` in \citet{maynez-etal-2020-faithfulness} such that we determine if the information in the summary is consistent with information in the source, and we do not consider `factuality' where valid external facts are acceptable. Existing unsupervised evaluation methods are categorized and explained in the following part.\footnote{Supervised approaches, with systems trained on human evaluation annotations, are outside the scope of this work.}

\subsection*{Textual overlap scores}

$n$-gram based metrics, including BLEU \cite {papineni-etal-2002-bleu}, ROUGE \cite{lin-2004-rouge}, and METEOR \cite{banerjee-lavie-2005-meteor} measure $n$-gram overlap between two texts. Instead of $n$-grams, BERTScore \cite{zhang2020bertscore} and BLEURT \cite{sellam-etal-2020-bleurt} compare texts in their representation space. These metrics measure {textual similarity}, so they are not necessarily a good measure of consistency. We note that the original works that proposed these metrics compare the summary against the ground-truth summary, but this work focuses on the scenario where there is no ground-truth summary, and these metrics are used as baselines to compare the summary against the source.

\subsection*{Knowledge representation}
\citet{triple_matching} assess factual consistency by comparing relation triples from the source and the summary. The relation triples are in the format of Subject-Relation-Object and can be obtained using a model-free method such as OpenIE \cite{etzioni2008open} or using a trained relation extraction model. The factual accuracy score based on the triple matching approach is then defined as,
\begin{equation*}
\text{Score} = \frac{|{\tt T}_{x} \cap {\tt T}_y|}{|{\tt T}_y|}
\end{equation*}
where ${\tt T}_x$ and ${\tt T}_y$ are relation triples extracted from the source and the summary, respectively. 

\subsection*{Textual Entailment}
Simulated data, such as real or fake summaries created by pre-defined transformations, have been used to train classifiers to detect inconsistent summaries \cite{kryscinski-etal-2020-evaluating, bao-etal-2022-suenes}. Alternatively, \cite{maynez-etal-2020-faithfulness} trained a textual entailment classifier on the Multi-NLI (MNLI) dataset \cite{williams-etal-2018-broad}. Given a context, the entailment model is to classify the hypothesis into one of the three classes (entail/neutral/contradict).  When applied to assess summaries, the context is the source document and the hypothesis is the summary. The probability of being the entail class is then used as the consistency score,
\begin{equation}
    \text{Score} = P(\text{entail}|x,y)
    \label{eq:entailment}
\end{equation}

\subsection*{Span-based Question Answering (SpanQAG)} 
A question-answering approach consists of a question-generation model and an answering model. Given automatically generated questions, the first answer is derived from the source and the second answer is derived from the evaluated summary, and then the two answers are compared. 

For example, \citet{eyal-etal-2019-question} proposed a QA-based method where questions are generated from the ground-truth summary. QAGS \cite{wang-etal-2020-asking} and FEQA \cite{durmus-etal-2020-feqa} generate questions from the evaluated summary, so these two methods are designed to measure the amount of information in the summary that is consistent with the source. In contrast, SummaQA \cite{scialom-etal-2019-answers} generates questions from the source document, so it assesses the coverage of the summary. As an extension to the ideas in QAGS/FEQA and SummQA, QuestEval \cite{scialom-etal-2021-questeval} generates questions from both the source and the summary separately to obtain a precision score and a recall score. QuestEval also assigns a weighting function to take into account the importance of each query/question.  

Nevertheless, existing QA methods are \textit{span-based} where the answering system extracts answer spans before two answer spans are compared. Due to the nature of span-based answers, answer verification (i.e. answer comparison) is typically through exact matching, token F1, BERTScore, or a learned metric \cite{deutsch-roth-2022-benchmarking}. This answer verification illustrates a drawback of the existing QA methods that they have to compare the similarity between two texts. To avoid span-based answer verification, we propose an alternative question answering-based approach where multiple-choice question generation and answering systems are used where the answers are now in the form of probability distributions rather than text spans. 

\section{Multiple-choice Question Answering and Generation (MQAG)}
\subsection{Motivation and Theory}
\label{section:proposed_method}
Since current summarization systems generate highly fluent summaries, this work focuses on assessing whether summaries contain the same information as that of the source, or whether it is contradictory. One way to view information would be to consider the set of questions that are answerable given a certain passage. If a summary is consistent with the source, then one would expect the set of answerable questions by the summary to overlap with those of the source and yield similar answers. Though span-based QA approaches are similarly motivated, existing span-based frameworks use text similarity measures, either in the form of lexical or representation space. In contrast, we attempt to measure information using multiple-choice questions, which allows for a more abstract understanding of information and enables convenient use of standard information-theoretic measures. 

\subsection{MQAG Score}
Let $x$ = source, $y$ = summary, $q$ = question, and $\mathbf{o}$ = options associated with the question $q$. We define information inconsistency as,
\begin{align}
    &\mathcal{I}(x,y) = \nonumber \\
    &\int\limits_{q,\mathbf{o}} D\left( P_{{\tt A}}(\mathbf{o}|q,x),P_{{\tt A}}(\mathbf{o}|q,y)   \right) P_{{\tt G}}(q,\mathbf{o}|y) \mathrm{d}\mathbf{o} \mathrm{d}q  \nonumber \\
    &\approx \frac{1}{N}\sum\limits_{i=1}^N D \left( P_{{\tt A}}(\mathbf{o}^{(i)}|q^{(i)},x), P_{{\tt A}}(\mathbf{o}^{(i)}|q^{(i)},y)   \right)
    \label{eq:mqag_equation}
\end{align}
where $\{q^{(i)}, \mathbf{o}^{(i)}\}$ is sampled from $P_{{\tt G}}(q,\mathbf{o}|y)$, the question-option generation model, $P_{{\tt A}}(\mathbf{o}^{(i)}|q^{(i)},x)$ and $P_{{\tt A}}(\mathbf{o}^{(i)}|q^{(i)},y)$ are the option distributions given the source and summary respectively, and $D$ is a statistical distance such as KL-divergence. Based on the information inconsistency score in Equation~\ref{eq:mqag_equation}, we define the MQAG score as,\footnote{If $D>1$, for example, when using KL-divergence, the MQAG score can be negative, but the maximum value is 1.0.} 
\begin{equation}
    \text{MQAG-Score}(x,y) = 1 - {\mathcal{I}}(x,y)
    \label{eq:mqag_sum}
\end{equation}
We refer to Equation~\ref{eq:mqag_sum} as the \textbf{MQAG-Sum} score as the questions are generated from the summary. Furthermore, it is possible to generate questions,  $\{q, \mathbf{o}\}$ using the source $x$ instead of the summary $y$, $\{q^{(i)}, \mathbf{o}^{(i)}\}$ is sampled from $P_{{\tt G}}(q,\mathbf{o}|x)$. We will refer to this variant as the \textbf{MQAG-Src} score. MQAG-Src is expected to measure the amount of source information present in the summary, i.e. the coverage of the summary, while MQAG-Sum is expected to measure the consistency of the summary with respect to the source. To account for consistency and coverage, we also consider a simple combination,
\begin{equation}
\text{MQAG-F1} = 2 \cdot \frac{\text{MQAG-Sum} \times \text{MQAG-Src}}{\text{MQAG-Sum} + \text{MQAG-Src}}
\label{eq:mqag_f1}
\end{equation}



\subsection{Statistical Distances $D$}
\label{section:distance}
Given two probability distributions over options $\mathbf{o}$ (e.g. one conditioned on source $x$, and the other conditioned on summary $y$), a statistical distance $D$ measures the distance between the probability distributions. There are multiple distances, which can be used, and in this work, we consider some of the main distances and investigate their properties as well as their empirical performance in our MQAG framework as follows,
\begin{itemize}
    \item KL-Divergence: 
    \begin{equation*}
        D_{\tt KL} = \sum_{o \in \mathbf{o}} P_{{\tt A}}({o}|q,x) \log \left( \frac{P_{{\tt A}}({o}|q,x)}{P_{{\tt A}}({o}|q,y)}  \right)
    \end{equation*}
    \item One-Best (i.e. argmax matching): 
    \begin{equation*}
        D_{\tt OB} = 
        \begin{cases}
        0,& \text{if } o_x =  o_y \\
        1,              & \text{otherwise}
        \end{cases}
    \end{equation*}
    where $o_x = \argmax_o P_{{\tt A}}({o}|q,x)$ and $o_y = \argmax_o P_{{\tt A}}({o}|q,y)$. $D_{\tt OB}$ simply determines whether the two answers match or not. 
    \item Total Variation: 
    \begin{equation*}
        D_{\tt TV} = \frac{1}{2} \left\Vert P_{{\tt A}}(\mathbf{o}|q,x) - P_{{\tt A}}(\mathbf{o}|q,y)  \right\Vert_1
    \end{equation*}
    \item Hellinger: 
    \begin{equation*}
        D_{\tt HL} = \frac{1}{\sqrt{2}} \left\Vert \sqrt{P_{{\tt A}}(\mathbf{o}|q,x)} - \sqrt{P_{{\tt A}}(\mathbf{o}|q,y)}  \right\Vert_2 
    \end{equation*}
\end{itemize}
KL divergence is unbounded, which means the value can be exceedingly large. In contrast, one-best is bounded but discontinuous. Both total variation and Hellinger distance are bounded and continuous. We illustrate examples of the properties of these statistical distances on Bernoulli distributions in Figure~\ref{fig:statistical_distances} in the appendix.

\section{Experimental Setup}

\subsection{System Development Data}
RACE \cite{lai-etal-2017-race} is a multiple-choice reading comprehension dataset where each example consists of context, question, answer, and 3 distractors (i.e. incorrect options). SQuAD \cite{rajpurkar-etal-2016-squad} is a collection of question-answer pairs derived from Wikipedia articles, and the correct answers can be any sequence of tokens in the given context. The statistics are provided in Table~\ref{tab:data_statistics_dev} where \textit{abstractiveness} is measured by 1.0 minus the length of the longest sequence that exists in both the context and the answer per the answer length, i.e. $1.0 - \text{ROUGE-L}_\text{Precision}(\text{Answer}, \text{Context})$.


\begin{table}[!ht]
\tabcolsep=1.1mm
  \centering
  \begin{tabular}{rcccc}
    \toprule
    \multirow{2}{*}{Dataset} &\multirow{2}{*}{Size} &\multicolumn{2}{c}{Length} &\multirow{2}{*}{Abstractive} \\
    & &Context  &Answer  \\
    \midrule
    SQuAD &98.2k  &317.8 &11.0 &0.0\% \\
    RACE  &97.7k  &138.3 &11.3 &39.1\% \\
    \bottomrule
  \end{tabular}
  \caption{Statistics of datasets for training MQAG systems. Length = \#words. Abstractiveness of 0\% indicates that in SQuAD the answer always exists in the context.}
  \label{tab:data_statistics_dev}
\end{table}

\subsection{Evaluation Data}
We evaluate the performance by measuring the correlation against human judgements at the summary level on QAG-(CNNDM \cite{cnndm_data}, XSum \cite{narayan-etal-2018-dont}), XSum-Hallucination and at the system level on Podcast Assessment and SummEval, and the definitions of summary-level and system-level correlations are provided in Appendix~\ref{appendix:correlation}. The statistics are provided in Table~\ref{tab:data_statistics}. 

\begin{table}[!ht]
\tabcolsep=1.2mm
  \centering
  \begin{tabular}{rccc}
    \toprule
    \multirow{2}{*}{Eval Dataset} &\multirow{2}{*}{Size} &\multicolumn{2}{c}{Length} \\
    & &Source  &Summary  \\
    \midrule     
    QAG-CNNDM  &235  &355.8 &54.4  \\
    QAG-XSum   &239  &403.7 &19.7  \\
    XSum-H     &2500 &442.1 &20.5  \\
    Podcast    &$^{*}$20$\times$179 &5950 &88.3 \\
    SummEval   &$^{*}$16$\times$100 &404.0 &63.7  \\
    \bottomrule
  \end{tabular}
  \caption{Statistics of evaluation datasets. Length is the number of words calculated using the NLTK tokenizer. $^{*}$\#systems$\times$documents.}
  \label{tab:data_statistics}
\end{table}

\noindent\textbf{QAG}. \citet{wang-etal-2020-asking} annotated 235 CNNDM summaries of the system in \citet{gehrmann-etal-2018-bottom} and 239 XSum summaries of fine-tuned BART \cite{lewis-etal-2020-bart}. The annotation was performed at the sentence level indicating if hallucination occurs or not. Subsequently, for each summary, the faithfulness (or consistency) score is then obtained by averaging all sentence-level human scores. 

\vspace{1.75mm}
\noindent\textbf{XSum-Hallucination (XSum-H)}.  \citet{maynez-etal-2020-faithfulness} annotated 2500 XSum summaries using 3 crowd-sourced workers on two metrics: 1) Faithfulness = whether the information is faithful w.r.t. the source at the token level. The judgements are then averaged; 2) Factuality = whether the summary level is factual w.r.t source and external knowledge. 

\vspace{1.75mm}
\noindent\textbf{Podcast Assessment}. \citet{manakul2022podcast} compiled 3580 podcast summaries of abstraction and extractive summarization systems from Spotify Podcast Challenge 2020 \cite{jones2021trec}. The human evaluation was performed on a 4-point scale considering a combination of consistency, coverage, and fluency. 

\vspace{1.75mm}
\noindent\textbf{SummEval}. \citet{fabbri-etal-2021-summeval} assessed 1600 CNNDM summaries from 16 different summarization systems on four aspects, including relevancy, consistency, coherency, and fluency. In this work, we use the consistency scores.

\subsection{Baselines}
All of the considered methods compare the summary $y$ against the source document $x$ without the ground-truth summary, and we implement these methods as described in Section~\ref{section:existing_method} using code/repository from the relevant previous works.

\vspace{1.75mm}
\noindent\textbf{ROUGE}. We use the ROUGE-1 (F1) score in the \texttt{rouge-score} Python package.

\vspace{1.75mm}
\noindent\textbf{OpenIE-TripleMatch}. The relation extraction is based on an open scheme, and we use the implementation in FactSumm \cite{factsumm}.

\vspace{1.75mm}
\noindent\textbf{BERTScore}. We use DeBERTa-base \cite{he2021deberta} fine-tuned to MNLI  as the backbone.

\vspace{1.75mm}
\noindent\textbf{Entailment model}. Following the method in \citet{maynez-etal-2020-faithfulness}, we trained BERT-large \cite{devlin-etal-2019-bert} on MNLI and we use the probability of the source being entailed by the summary as the assessment score as shown in Equation~\ref{eq:entailment}. 

\vspace{1.75mm}
\noindent\textbf{Span-based QAG Baselines}. We use three existing span-based question-answering methods as our baselines: {QAGS} proposed by \citet{wang-etal-2020-asking}, FEQA proposed by \citet{durmus-etal-2020-feqa}, and QuestEval proposed by \citet{scialom-etal-2021-questeval}.



\subsection{MQAG Implementation}
\subsubsection*{Question Generation ({\tt G1}, {\tt G2})}
The multiple-choice question generation is implemented in two stages.\footnote{The motivation is based on our initial experiments that a single generation system (generating the question and 4 options together) often gave low-quality distractors, and using two generation systems improved the quality of distractors.} First model ${\tt G1}$ generates the question $q$ and answer $a$, then model ${\tt G2}$ generates the distractors $\mathbf{o}_{\setminus a}$ given $q$ and $a$.
\begin{equation}
    P_{{\tt G}}(q,\mathbf{o}|y) = P_{{\tt G2}}(\mathbf{o}_{\setminus a}|q,a,y) P_{{\tt G1}}(q,a|y) 
\end{equation}
where $\mathbf{o} = \{a, \mathbf{o}_{\setminus a}\}$ denotes all options/choices. We set the number of options to four. Both {\tt G1} and {\tt G2} are sequence-to-sequence T5-large models \cite{raffel2020exploring}. The question-answer generation system {\tt G1} is fine-tuned to either RACE or SQuAD, and the distractor generation system {\tt G2} is fine-tuned to RACE.

\subsubsection*{Question Answering ({\tt A})}
The answering stage contains one model ${\tt A}$, which is Longformer-large \cite{beltagy2020longformer} with a multiple-choice setup following \citet{Yu2020ReClor, raina-gales-2022-answer}. The input to the model is a concatenation of context, question and option. The answering model ${\tt A}$ is fine-tuned to RACE.


\subsubsection*{Answerability of Generated Questions}
Because not all generated questions are of high quality, we consider filtering out low-quality questions through question-context answerability measures \cite{kundu-ng-2018-nil, hu2019read}. We consider a simple answerability measure based on the entropy of the probability distribution over the options. We define the effective number of options,
\begin{equation}
    \mathcal{N}_y(q, \mathbf{o}) = 2 ^ {\mathcal{H} [ P_{{\tt A}}(\mathbf{o}|q,y) ]}
    \label{eq:answerability_entropy}
\end{equation}
where $\mathcal{H}(.)$ is base-2 entropy, so $\mathcal{N}_y(q, \mathbf{o})$ ranges from 1.0 to the number of options, e.g. 4.0. When $q$ is generated from $y$ but $\mathcal{N}_y(q, \mathbf{o})$ is high, this question $q$ should be deemed \textit{unanswerable} as it is not answerable even when using the same context. As a result, we use $\mathcal{N}_y(q, \mathbf{o})$ as an answerability criterion to \textit{reject} questions which have $\mathcal{N}_y(q, \mathbf{o})$ higher than a threshold denoted by $\mathcal{N}^{\tau}_y$. 


\section{Experimental Results}

\subsection{Analysis of the Components in MQAG}
In this subsection, we carry out experiments to find the best configuration of MQAG, including the analysis of statistical distances, variants of MQAG, and answerability. We build two MQAG variants: MQAG$_\text{SQuAD}$ and MQAG$_\text{RACE}$, which differ in the training data of the question+answer generator {\tt G1}, while the distractor generator {\tt G2} and answering system {\tt A} are both trained on RACE. 

\begin{figure*}[!ht]
    \centering
\includegraphics[width=\linewidth,keepaspectratio]{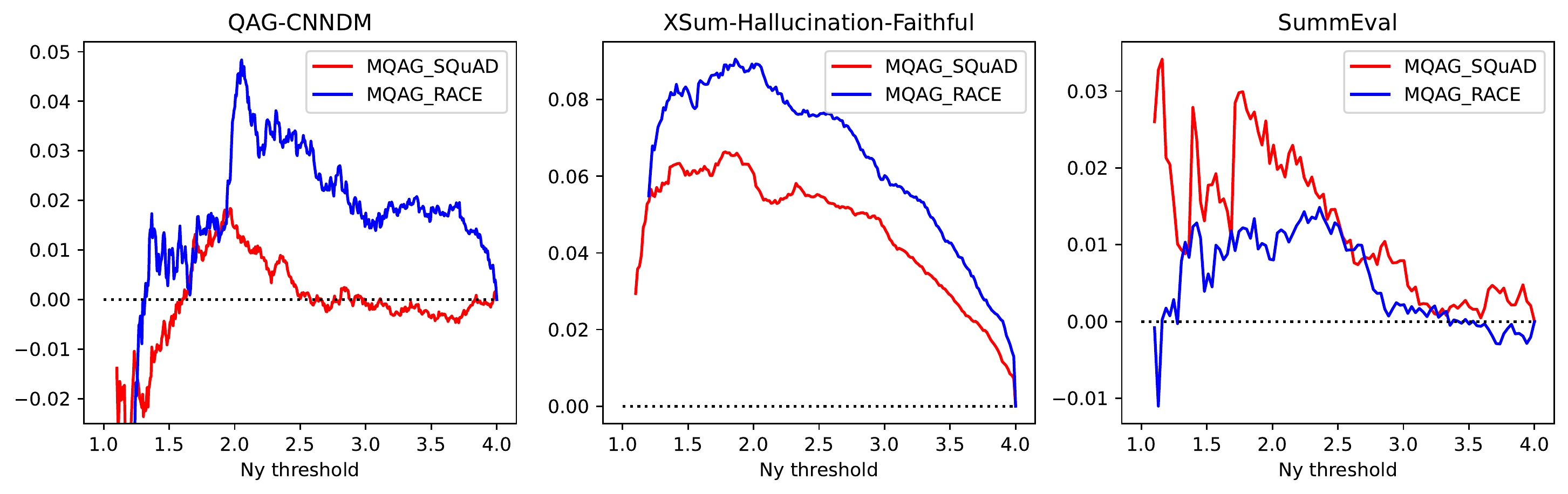}
    \caption{$\Delta$PCC of MQAG-Sum with total variation (i.e. PCC $-$ PCC$_{N^{\tau}_y = 4.0}$) against the answerability threshold $\mathcal{N}^{\tau}_y$ on X-axis. MQAG without answerability is equivalent to setting ${N^{\tau}_y = 4.0}$, and the results at this operating point can be seen on the right-most point in each plot. As we reduce the threshold ($N^{\tau}_y\downarrow$), more questions are rejected. The results on QAG-XSum and Podcast are provided in Figure~\ref{fig:answerability_appendix} in the appendix.}
    \label{fig:answerability}    
\end{figure*}

\subsubsection*{Statistical Distances}
In Table~\ref{tab:results_distance}, our results compare statistical distances. It can be seen that in both configurations, KL-divergence yields lower correlations than other distances, and on average total variation slightly outperforms Hellinger and one-best distances. Hence, total variation will be used as the main distance. The next observation is that MQAG$_\text{SQuAD}$, despite generating more extractive questions, achieves higher correlations than MQAG$_\text{RACE}$ on most tasks except on Podcast and SummEval.

\begin{table}[!ht]
\tabcolsep=1.9mm
  \small
  \centering
  \begin{tabular}{ccccccc}
    \toprule
    \multirow{2}{*}{$D$} &\multicolumn{2}{c}{QAG} &\multicolumn{2}{c}{XSum-H} &\multirow{2}{*}{Podc} &\multirow{2}{*}{SumE} \\
    &CNN &XSum &Faith &Fact  & & \\
    \midrule
    \rowcolor{Gray}
    \multicolumn{7}{l}{MQAG-Sum, {\tt G1} = SQuAD} \\
    $D_{\tt KL}$ &0.478 &0.374 &0.177 &0.226 &0.251 &0.936 \\
    $D_{\tt OB}$ &0.476 &0.354 &0.295 &0.254 &0.677 &0.872  \\
    $D_{\tt TV}$ &0.508 &0.396 &0.269 &0.267 &0.225 &0.870  \\
    $D_{\tt HL}$ &0.499 &0.399 &0.266 &0.269 &0.201 &0.870  \\
    \rowcolor{Gray}
    \multicolumn{7}{l}{MQAG-Sum, {\tt G1} = RACE} \\
    $D_{\tt KL}$ &0.450 &0.283 &0.135 &0.179 &0.789 &0.954\\
    $D_{\tt OB}$ &0.453 &0.225 &0.240 &0.221 &0.839 &0.928\\
    $D_{\tt TV}$ &0.462 &0.309 &0.221 &0.244 &0.770 &0.933\\
    $D_{\tt HL}$ &0.473 &0.323 &0.215 &0.244 &0.751 &0.927\\
    \bottomrule
  \end{tabular}
  \caption{Comparison of Statistical Distances using MQAG-Sum without answerability.}
  \label{tab:results_distance}
\end{table}

\subsubsection*{MQAG-Sum, MQAG-Src, MQAG-F1}
Here, we compare three variants of MQAG scores. Our results in Table~\ref{tab:results_tv_race_src_sum_f1} show that {MQAG-Src}, which assesses how much source information is contained in the summary by generating questions from the source, achieves lower PCCs than MQAG-Sum on all datasets. This finding aligns with our expectation, as the summaries were graded by humans predominantly on the consistency aspect (which MQAG-Sum was designed to measure) rather than the quantity of source information present (which MQAG-Src measures). When combining MQAG-Src and MQAG-Sum into MQAG-F1, we only observe a small gain on two test settings. Therefore, MQAG-Sum is selected as our main MQAG configuration for the remaining investigations.

\begin{table}[!ht]
\tabcolsep=1.9mm
  \centering
    \small
  \begin{tabular}{cccccccc}
    \toprule
    \multirow{2}{*}{} &\multicolumn{2}{c}{QAG} &\multicolumn{2}{c}{XSum-H} &\multirow{2}{*}{Podc} &\multirow{2}{*}{SumE} \\
    &CNN &XSum &Faith &Fact  & & \\
    \midrule
    \rowcolor{Gray}
    \multicolumn{7}{l}{{\tt G1} = SQuAD, $D$ = Total Variation} \\
    Sum &0.508 &0.396 &0.269 &0.267 &0.225 &0.870  \\
    Src &0.272 &0.017 &0.093 &0.037 &0.470 &0.707 \\
    F1  &0.490 &0.393 &0.286 &0.261 &0.475 &0.863 \\
    \rowcolor{Gray}
    \multicolumn{7}{l}{{\tt G1} = RACE, $D$ = Total Variation} \\
    Sum &0.462 &0.309 &0.221 &0.244 &0.770 &0.933 \\
    Src &0.233 &0.143 &0.069 &0.087 &0.144 &0.588 \\
    F1  &0.468 &0.301 &0.217 &0.252 &0.731 &0.866 \\
    \bottomrule
  \end{tabular}
  \caption{Comparison of MQAG-Src, MQAG-Sum, and MQAG-F1 without answerability.}
  \label{tab:results_tv_race_src_sum_f1}
\end{table}

\subsubsection*{Answerability}
In Figure~\ref{fig:answerability}, the answerability is swept from 4.0 (keeping all questions) to 1.0 (only keeping those that the answering system {\tt A} is highly confident). It can be seen that as we filter out high-entropy questions, there is an upward trend in performance across all tasks. In addition, as shown in the figure, setting $\mathcal{N}^{\tau}_y$ at 2.0 seems to be a reasonable answerability threshold. At this threshold, $\mathcal{N}^{\tau}_y=2.0$, out of 50 automatically generated questions, about 36 questions are kept for MQAG$_\text{SQuAD}$ and about 30 questions are kept for MQAG$_\text{RACE}$. The number of remaining questions is similar across all datasets as shown in Table~\ref{tab:num_kept} in the appendix. Thus, we set $\mathcal{N}^{\tau}_y=2.0$, and the performance of MQAG using this answerability criterion is presented and compared against baseline systems in Table~\ref{tab:baseline_vs_mqag}.




\subsection{Comparison Against Existing Baselines}
\begin{table*}[!t]
  \centering
\scalebox{0.96}{
  \begin{tabular}{lcccccc}
    \toprule
    \multirow{2}{*}{Method} &\multicolumn{2}{c}{QAG} &\multicolumn{2}{c}{XSum-H} &\multirow{2}{*}{Podcast} &\multirow{2}{*}{SumEvl} \\
     &CNNDM &XSum &Faithful &Factual  &  \\
    \midrule
    \rowcolor{Gray}
    \multicolumn{7}{l}{\textit{Baselines: Other Approaches}} \\
    ROUGE-1                  &0.337  &0.012 &-0.050 &0.008  &0.326  &0.458  \\
    OpenIE-TripleMatching    &0.381  &0.131 &0.019  &-0.020 &0.706  &0.548  \\
    BERTScore                &0.584  &0.008 &0.185  &0.154  &0.718  &0.645  \\
    Entailment (BERT Model)  &0.159  &0.169 &0.362  &0.209  &0.228  &0.619  \\
    \rowcolor{Gray}
    \multicolumn{7}{l}{\textit{Baselines: SpanQAG}} \\
    QAGS                     &0.437  &0.200 &0.101  &0.080  &0.464  &0.812  \\
    FEQA                     &0.322  &0.283 &0.297  &0.171  &0.603  &0.464  \\
    QuestEval                &0.250  &0.173 &0.421  &0.197  &0.579  &0.838  \\
    \rowcolor{Gray}
    \multicolumn{7}{l}{\textbf{Multiple-choice Question Answering and Generation (MQAG)}} \\
    MQAG$_\text{SQuAD}$   &\underline{0.519} &\underline{0.407} &0.324 &\underline{0.292} &0.502 &\underline{0.890} \\
    MQAG$_\text{RACE}$    &\underline{0.502} &\underline{0.313} &0.306 &\underline{0.270} &\underline{0.855} &\underline{0.945} \\
    \bottomrule
  \end{tabular}}
  \caption{Pearson Correlation Coefficient (PCC) between the scores of summary evaluation methods and human judgements. PCCs are computed at the summary level on QAG and XSum-H, and at the system level on Podcast and SummEval. PCCs on Podcast are computed on 15 abstractive systems. Our best performing MQAG configuration consists of (i) generation stage {\tt G} generates questions from summary $y$ (i.e. MQAG-Sum), (ii) statistical distance is total variation, (iii) the answerability threshold $\mathcal{N}^{\tau}_y$ is set to 2.0. Underline denotes where MQAG outperforms the best SpanQAG system, which is 5 out of 6 tasks. When compared to all baselines, MQAG achieves the highest PCC on 4 out of 6 tasks. The results of all MQAG configurations are provided in Table~\ref{tab:mqag_pearson_appendix}, and Spearman's correlation results are provided in Table~\ref{tab:baseline_vs_mqag_spearman} in the appendix.}
  \label{tab:baseline_vs_mqag}
\end{table*}

\noindent The baseline and MQAG results are shown in Table~\ref{tab:baseline_vs_mqag}. The observation is that MQAG achieves a higher correlation than the best SpanQAG on 5 out of 6 tasks. When compared to all existing baselines, MQAG achieves state-of-the-art performance on 4 out of 6 tasks. To investigate the impact of the abstractiveness of summaries on the performance, we split QAG-XSum and XSum-H datasets\footnote{XSum summaries are more abstractive than CNNDM summaries, so using XSum should enable us to investigate the impact of abstractiveness better than CNNDM.} into two portions of the same size by abstractiveness as measured by the longest sequence in the summary that exists in the source per the summary length (i.e. ROUGE-L precision of summary $y$ using source $x$ as the reference). The results in Table~\ref{tab:results_by_abstractiveness} show that although MQAG$_\text{RACE}$ achieves lower PCCs than MQAG$_\text{SQuAD}$ (in Table~\ref{tab:baseline_vs_mqag}), when evaluated on the more abstractive split, the performance MQAG$_\text{RACE}$ is much closer to that of MQAG$_\text{SQuAD}$. In addition, compared to MQAG, SpanQAG methods show a larger drop in PCCs in the more abstractive split. This finding further illustrates the benefits of comparing answer distributions rather than text spans.

\begin{table}[!ht]
  \centering
  \begin{tabular}{lcccc}
    \toprule
    \multirow{2}{*}{Method} &\multicolumn{2}{c}{QAG-XSum} &\multicolumn{2}{c}{XSum-H} \\
    &Low &High &Low &High \\
    \midrule
    QAGS     &0.190 &0.184 &0.101 &0.159 \\
    FEQA     &0.296 &0.163 &0.290 &0.124 \\
    QuestEval&0.215 &0.061 &0.398 &0.326 \\
    \midrule
    MQAG$_\text{SQuAD}$ &0.431 &0.328 &0.334 &0.254 \\
    MQAG$_\text{RACE}$  &0.277 &0.295 &0.319 &0.249 \\
    \bottomrule
  \end{tabular}
  \caption{Performance as measured by Pearson correlation coefficient on the \textit{low} abstractiveness and \textit{high} abstractiveness of QAG-XSum and XSum-H (Faithful). The results on the entire datasets are in Table~\ref{tab:baseline_vs_mqag}.}
  \label{tab:results_by_abstractiveness}
\end{table}

\begin{figure*}[!ht]
    \centering
\includegraphics[width=0.9\linewidth,keepaspectratio]{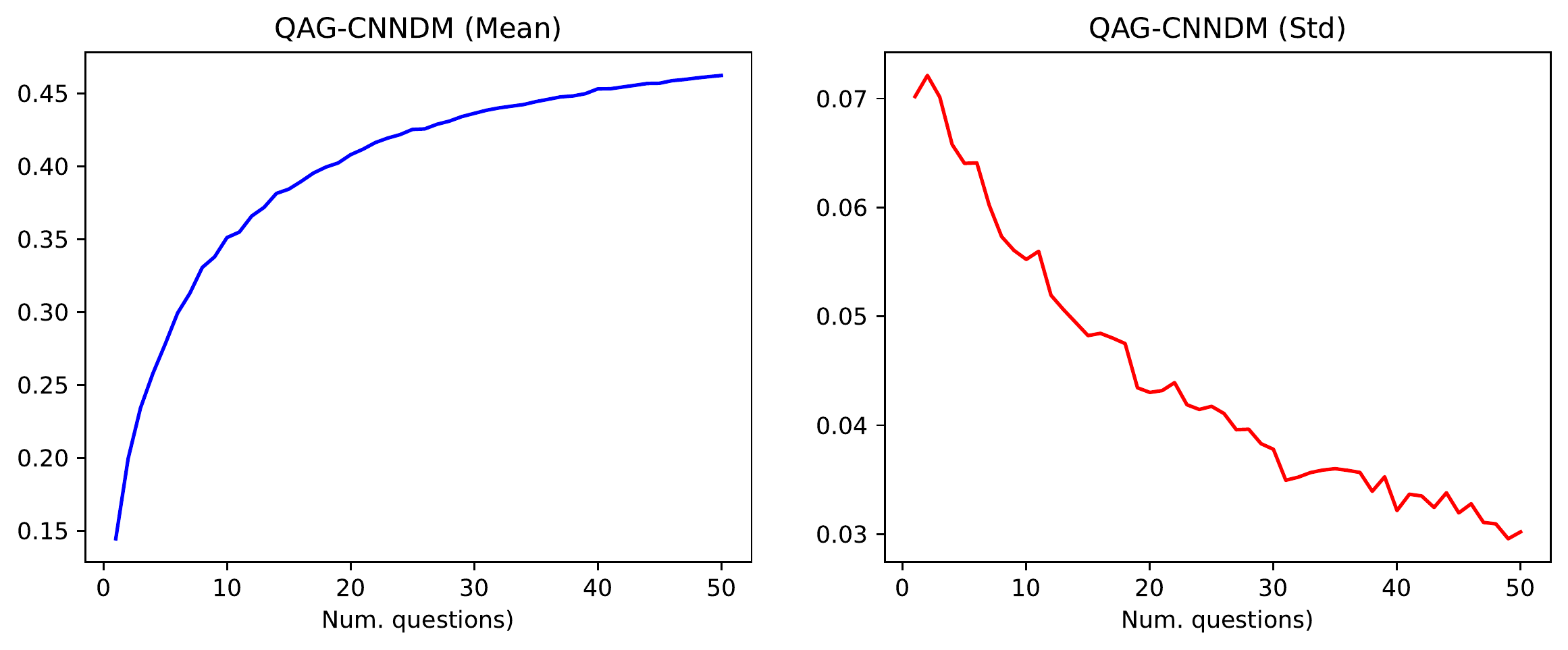}
    \caption{Mean and standard deviation of Pearson correlation (Y-axis) of MQAG$_\text{RACE}$ on QAG-CNNDM when the number of generated questions $N$ is varied from 1 to 50 (X-axis). Standard deviation is obtained via bootstrapping. The results on other datasets are provided in Figure~\ref{fig:numQ_appendix} in the appendix.}
    \label{fig:numQ}    
\end{figure*}

\section{Ablation Studies}
\subsection{Number of Questions ($N$)}
We analyse the impact of the number of generated questions on the performance of MQAG. The mean and standard deviation are presented in Figure~\ref{fig:numQ}. The results show a smooth increase in correlation, which is as expected because the framework is based on a Monte-Carlo approximation (in Equation~\ref{eq:mqag_equation}), and a similar finding was also observed in QAGS \cite{wang-etal-2020-asking}. Figure~\ref{fig:numQ} also shows that the variance decreases with $N$, showing the stability of the approach. Though the performance curve has not completely plateaued at $N$=50, since the computational cost of MQAG scales linearly with $N$, 50 questions seem to be a reasonable compromise between computational efficiency and performance. An interesting next step would be to investigate if the same or similar performance can be achieved with as low $N$ as possible, for example, by generating a smaller but more diverse set of questions and options such as varifocal question generation where questions are generated based on different focal points \cite{ousidhoum-etal-2022-varifocal}.

\subsection{Model Choices}
\subsubsection*{Pre-trained Backbone}
We investigate model choices by swapping to less capable models, e.g. T5-large $\rightarrow$ T5-base for generation, and Longformer(4096) $\rightarrow$ RoBERTa(512) \cite{liu2019roberta} for answering. The results in Table~\ref{tab:ablation_models} in the appendix show: (1) For generation stage, using a smaller model does not result in lower performance. This could be because T5-base has higher perplexity, and yields more diverse questions. (2) In contrast, for answering stage, when using RoBERTa, with a shorter input length, the performance on SummEval (the input length is mostly shorter than 512) remains almost the same. However, as the input length is longer in other datasets, we observe a drop in PCC when using RoBERTa.  

\subsubsection*{Zero-shot Multiple-choice Question Generation}
Given the impressive results of large language models (LLMs) across natural language generation tasks, we investigate the performance of LLMs in a zero-shot fashion instead of using fine-tuned T5 for multiple-choice question generation. Specifically, we use OpenAI GPT-3 \cite{gpt3brown} (text-davinci-003) where we query 50 questions and 4 options using the following prompt format: \vspace{2.25mm} \\
{\tt
Write 50 diverse multiple-choice questions with 4 options from the following context:\ \{context\}.}
\vspace{2.25mm}

\noindent We found that GPT-3 generated 50 questions as specified in the prompt around 26\% of the examples and the remaining only have 20 questions. The majority of questions (more than 95\%) have 4 options, while the remaining have 2 options. In Table~\ref{tab:gpt3_results}, the results show that zero-shot GPT-3 performs worse than our fine-tuned T5 systems in both multiple-choice question generation tasks. This illustrates that there is some sensitivity due to the quality of generated questions, and using our fine-tuned T5 is a better option than zero-shot GPT-3.

\begin{table}[!ht]
  \centering
  \begin{tabular}{ccc}
    \toprule
    \multirow{2}{*}{Backbone} &\multicolumn{2}{c}{QAG} \\
    &CNNDM &XSum \\
    \midrule
    T5 (SQuAD) &0.508 &0.396 \\
    T5 (RACE) &0.462 &0.309 \\ 
    GPT-3   &0.392 &0.130 \\  
  \bottomrule
  \end{tabular}
  \caption{GPT-3 versus fine-tuned T5 using $D_{\tt TV}$ without answerability for multiple-choice question generation.}
  \label{tab:gpt3_results}
\end{table}

\section{Conclusion}
This work proposes MQAG -- a novel scheme for assessing information consistency between source and summary based on the distance between multiple-choice answer distributions instead of text-based answer spans in existing question-answering methods. Our experiments demonstrate the potential of this alternative approach which outperforms existing techniques on various datasets. The realization of the framework exploits current multiple-choice question generation and answering systems. Its performance is expected to increase as backbone systems improve, for example, the diversity of questions generated and the selection of options. Also, the framework is highly interpretable, allowing more insight into summary assessment.


\newpage
\section*{Limitations}
\textit{Domain}. Our approach is designed to assess the information content, so it may not work well with other aspects of summary evaluation such as fluency or coherency. Our analysis is based on the systems trained on RACE, which is collected from English examinations in China. Hence, the generated questions and answer distributions could be biased towards the style of the examinations. 

\vspace{1.75mm}
\noindent\textit{Efficiency}. Given the realization of the MQAG framework where two generators {\tt G1} and {\tt G2} are adopted, the MQAG framework can be slow when using old infrastructure, for example, it takes around 3 seconds per question on one NVIDIA P100 GPU. To address this issue, future work could explore a more efficient realization of MQAG.



\section*{Acknowledgments}
This work is supported by Cambridge University Press \& Assessment (CUP\&A), a department of The Chancellor, Masters, and Scholars of the University of Cambridge, and the Cambridge Commonwealth, European \& International Trust. We would like to thank the anonymous
reviewers for their helpful comments.

\bibliography{anthology,custom}
\bibliographystyle{acl_natbib}

\appendix
\section{More Details about Models and Data}
\subsection*{Training QG and QA systems}
We train the question+answer generation model ({\tt G1}) on RACE or SQuAD, and train the distractor generation model ({\tt G2}) and the answering model ({\tt A}) on RACE. We do early stopping when the performance on the validation set does not improve. We use batch size 8 for {\tt G1} and {\tt G2} models (T5) and 2 for {\tt A} model (Longformer). The learning rate is set to 1e-6, and we use the Adam optimizer. We carried out training on one NVIDIA A100-80GB GPU. Training one generation model (T5-large) takes around 8 hours, and training the answering model (Longformer-4096) takes up to 2 days. Running MQAG inference with generation=T5-large and answering=Longformer-4096 on one NVIDIA P100 GPU takes around 3 seconds per question.

\subsection*{Licenses}
The licenses of the datasets are CC-BY-4.0 for XSum-Hallucination and Podcast Assessment, and MIT license for SummEval. For QAG, we were unable to find its license. The licenses of T5 and Longformer backbone models are apache-2.0.

\subsection*{Open-Sourcing Trained Models}
To allow the trained models in MQAG to be used for \textit{research} purposes in other question generation and answering tasks, we have made them available online. The links to these models on HuggingFace can be found on our project page at \url{https://github.com/potsawee/mqag0}.

\section{Statistical Distances}
\begin{figure}[!h]
    \centering
\includegraphics[width=\linewidth,keepaspectratio]{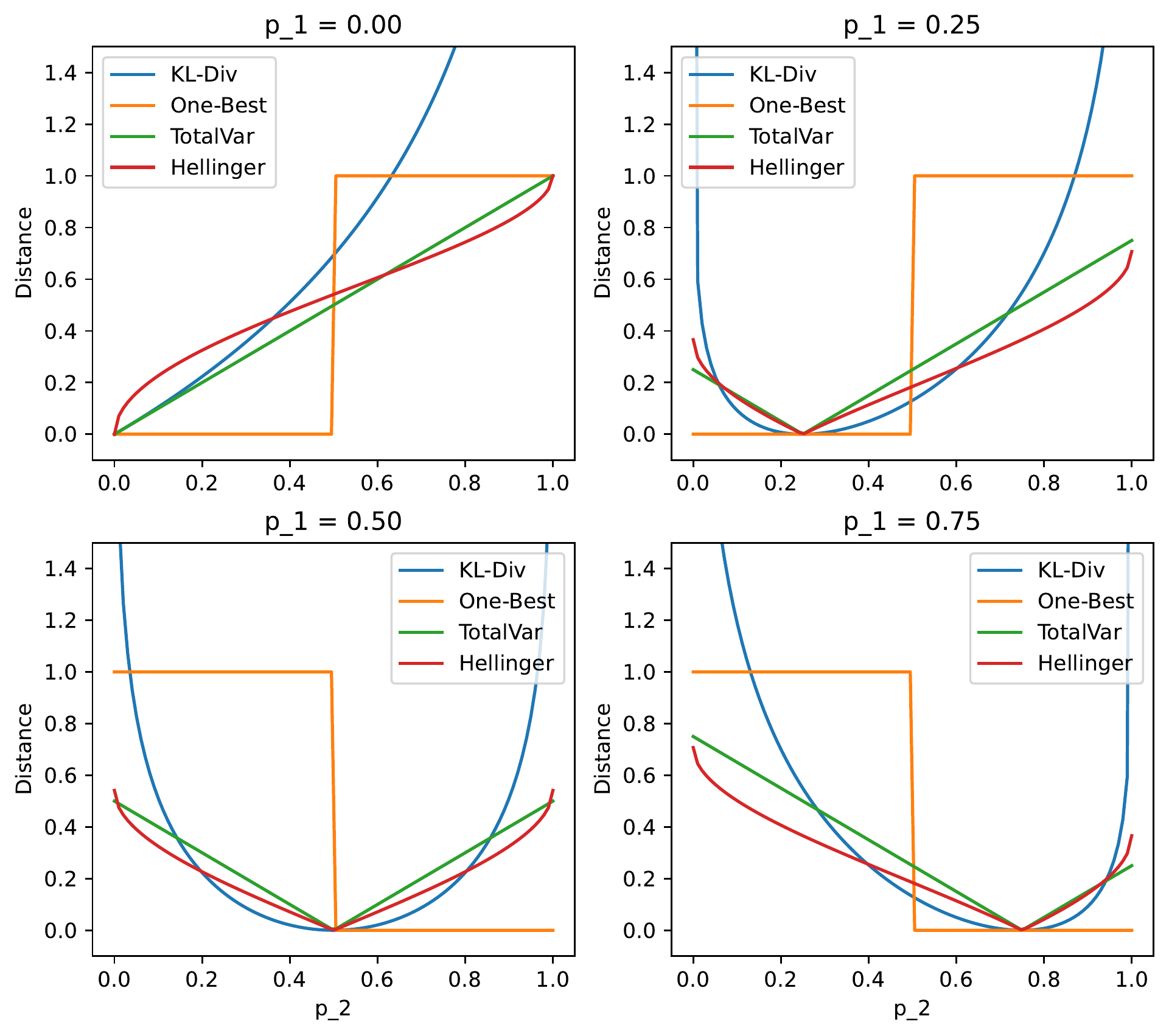}
    \caption{Statistical distances between two Bernoulli distributions $\mathbf{p}_1 = [p_1; 1-p_1]$ and $\mathbf{p}_2 = [p_2; 1-p_2]$ at different values of $p_1$. We show 4 plots of different values of $p_1$ = 0.00, 0.25, 0.50, 0.75, and Y-axis represents distance $D$ and X-axis represents $p_2$. It can be seen that KL divergence is unbounded, which means the value can be exceedingly large. One-best, in contrast, is bounded between 0.0 and 1.0; however, one-best is discontinuous. Total variation and Hellinger distance are continuous and bounded between 0.0 and 1.0.}
    \label{fig:statistical_distances}    
\end{figure}

\section{Computing Correlation}
\label{appendix:correlation}
Following the notation in \citet{deutsch-etal-2021-towards}, let $z_i^j$ and $\bar{z}_i^{j}$ be two scores of metrics $Z$ and $\bar{Z}$ for the summary output by system $i \in \{1,...,N\}$ on the document $j \in \{1,...,M\}$. In this work, $Z$ is the evaluation method, and $\bar{Z}$ is the human judgement. The correlations, e.g. Pearson or Spearman's rank correlation coefficient, are defined as follows:
\begin{itemize}
    \item System-level ({i.e.} Corpus-level)
    \begin{equation*}
        \rho = \text{Corr} \left( \left\{ \frac{\sum_j z_i^j}{M} , \frac{\sum_j \bar{z}_i^j}{M}\right\}_{i=1}^N \right)
    \end{equation*}
    \item Summary-level ({i.e.} Sentence-level)
    \begin{equation*}
        \rho = \frac{1}{M} \sum_j \text{Corr} \left( \left\{z_i^j, \bar{z}_i^j\right\}_{i=1}^N\right)
    \end{equation*}
\end{itemize}

\section{Additional Results}
\subsection{Ablation: Model Choices}
For generation models, we measure cross-entropy losses on RACE-testset:
\begin{itemize}
    \item T5-base (223M):   {\tt G1} = 1.612,   {\tt G2} = 1.875
    \item T5-large (738M):  {\tt G1} = 1.478,   {\tt G2} = 1.741
\end{itemize}
where {\tt G1} denotes question+answer generation, and {\tt G2} denotes distractor generation. For answering models, we measure accuracy on RACE-testset:
\begin{itemize}
    \item Roberta (355M):   Accuracy = 84.84
    \item Longformer (435M): Accuracy = 81.67
\end{itemize}

\begin{table}[!ht]
\tabcolsep=0.07cm
  \centering
  \begin{tabular}{ccccc}
    \toprule
    \multicolumn{2}{c}{Model}  &\multicolumn{3}{c}{Pearson Corr.} \\
    Generation &Answering  &SumE &QAG-X &Podc  \\
    \midrule     
    T5-base     &RoBERTa    &0.949 &0.242 &0.471 \\
    T5-base     &Longformer &0.949 &0.293 &0.647 \\
    T5-large    &RoBERTa    &0.930 &0.211 &0.350 \\
    T5-large    &Longformer &0.930 &0.229 &0.772 \\
    \bottomrule
  \end{tabular}
  \caption{Ablation on model choices in MQAG using $N$=20. SumE = SummEval (Consistency aspect), QAG-X = QAG-XSum, Podc = Podcast Assessment.}
  \label{tab:ablation_models}
\end{table}

\subsection{MQAG Results}
\begin{figure*}[!ht]
    \centering
\includegraphics[width=0.7\linewidth,keepaspectratio]{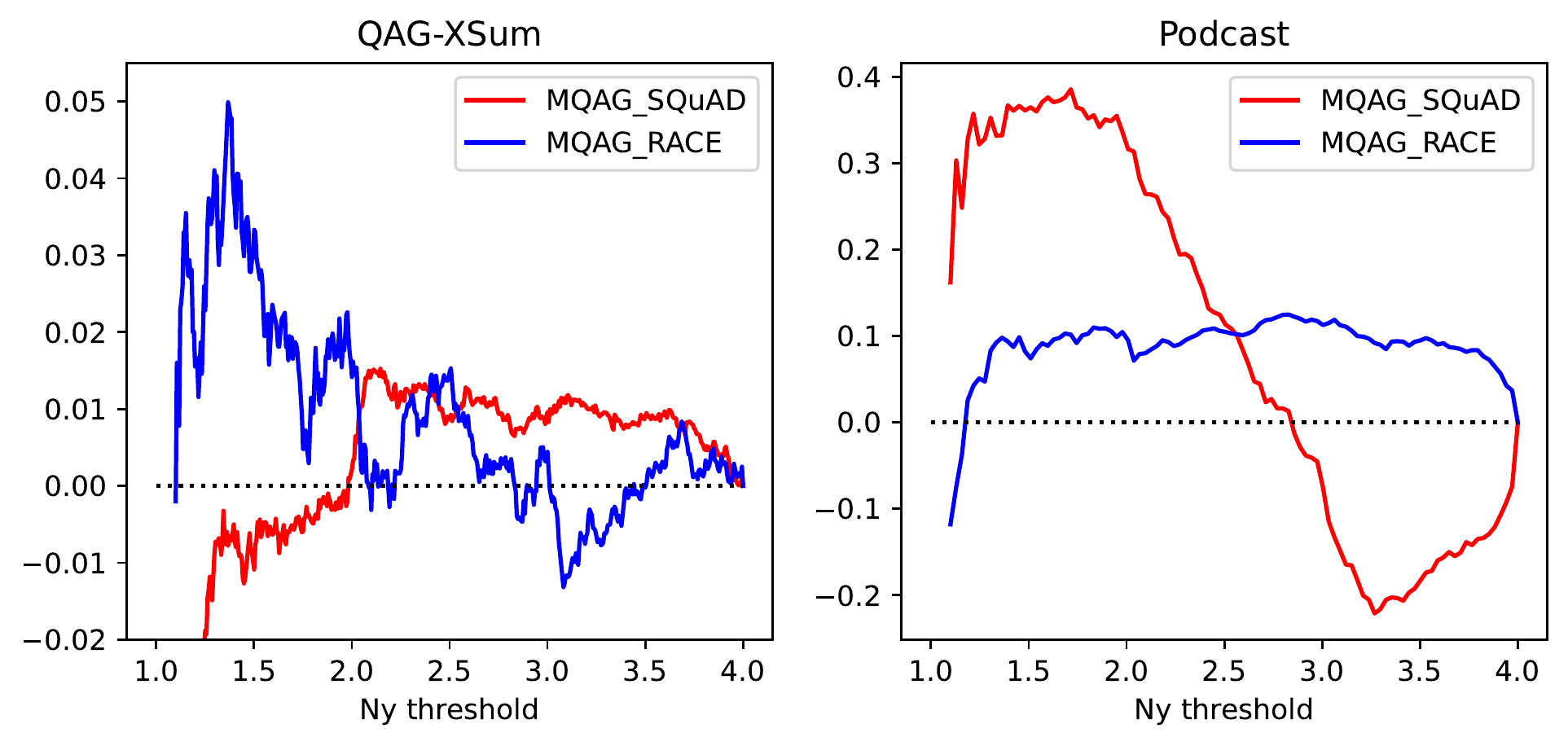}
    \caption{$\Delta$PCC of MQAG-Sum with total variation against the answerability threshold $\mathcal{N}^{\tau}_y$ on the X-axis. This figure extends Figure~\ref{fig:answerability} in the main text.}
    \label{fig:answerability_appendix} 
\end{figure*}

\begin{figure*}[!ht]
    \centering
\includegraphics[width=\linewidth,keepaspectratio]{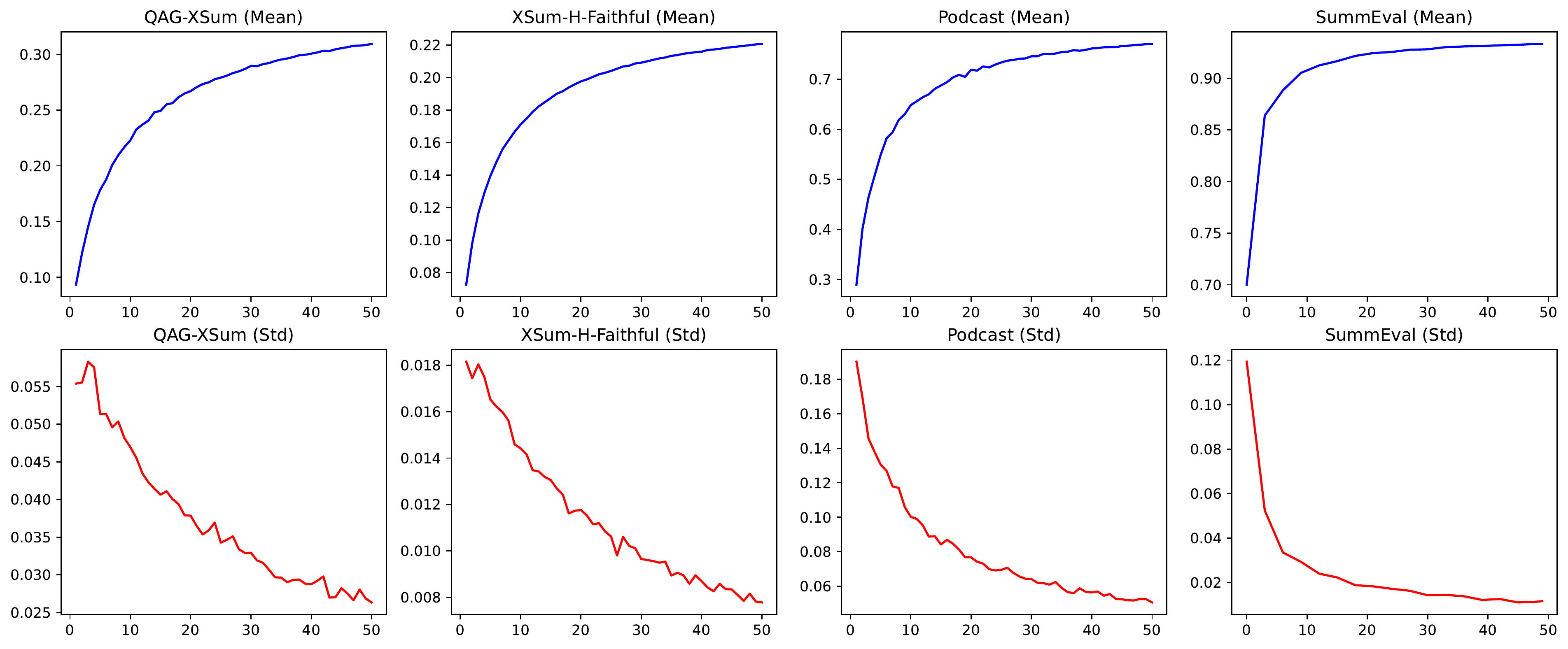}
    \caption{Mean (top row) and standard deviation (bottom row) of Pearson correlation (Y-axis) of MQAG$_\text{RACE}$ when the number of generated questions $N$ is varied from 1 to 50 (X-axis). This figure extends Figure~\ref{fig:numQ} in the main text.}
    \label{fig:numQ_appendix}    
\end{figure*}

Here, we provide results that are complementary to those presented in the main text. Figure~\ref{fig:answerability_appendix} illustrates the answerability results on QAG-XSum and Podcast, and Figure~\ref{fig:numQ_appendix} illustrates the impact of $N$ on the remaining datasets not presented in the main text. Table~\ref{tab:mqag_pearson_appendix} shows the results of all MQAG configurations. Table~\ref{tab:baseline_vs_mqag_spearman} shows the Spearman's rank correlation coefficient of the main results.

\begin{table*}[!ht]
  \centering
  \begin{tabular}{cccccc}
    \toprule
    Method  &QAG-CNNDM &QAG-XSum &XSum-H &Podcast &SummEval \\
    \midrule
    MQAG$_\text{SQuAD}$ &35.0 &37.4 &34.0 &34.7 &37.0 \\
    MQAG$_\text{RACE}$  &30.5 &30.0 &30.0 &30.5 &31.1 \\
    \bottomrule
  \end{tabular}
  \caption{The number remaining questions at $\mathcal{N}^{\tau}_y=2.0$.}
  \label{tab:num_kept}
\end{table*}

\begin{table*}[!t]
\tabcolsep=2mm
  \centering
  \begin{tabular}{ccccccccccc}
    \toprule
    \multicolumn{4}{c}{\textbf{MQAG Configuration}} &\multicolumn{2}{c}{QAG} &\multicolumn{2}{c}{XSum-H} &\multirow{2}{*}{Podcast} &\multirow{2}{*}{SumEvl} \\
{\tt G}'s Inp. &{\tt G1}-trained &Dist.  &Ans.  &CNNDM &XSum &Faithful &Factual  & & \\
    \midrule
    Src $x$&SQuAD &$D_{\tt KL}$  &\xmark       &0.219 &0.008 &0.070 &0.027 &0.432 &0.726 \\
    Src $x$&SQuAD &$D_{\tt OB}$  &\xmark       &0.264 &0.003 &0.165 &0.064 &0.788 &0.703 \\
    Src $x$&SQuAD &$D_{\tt TV}$  &\xmark       &0.272 &0.017 &0.093 &0.037 &0.470 &0.707 \\
    Src $x$&SQuAD &$D_{\tt HL}$  &\xmark       &0.266 &0.010 &0.081 &0.032 &0.517 &0.713 \\
    \hdashline
    Sum $y$&SQuAD &$D_{\tt KL}$&\xmark         &0.478 &0.374 &0.177 &0.226 &0.251 &0.936 \\
    Sum $y$&SQuAD &$D_{\tt OB}$&\xmark         &0.476 &0.354 &0.295 &0.254 &0.677 &0.872  \\
    Sum $y$&SQuAD &$D_{\tt TV}$&\xmark         &0.508 &0.396 &0.269 &0.267 &0.225 &0.870  \\
    Sum $y$&SQuAD &$D_{\tt HL}$&\xmark         &0.499 &0.399 &0.266 &0.269 &0.201 &0.870  \\
    \hdashline
    F1  &SQuAD &$D_{\tt KL}$  &\xmark          &0.508 &0.361 &0.197 &0.213 &0.531 &0.921 \\
    F1  &SQuAD &$D_{\tt OB}$  &\xmark          &0.416 &0.161 &0.296 &0.199 &0.825 &0.869 \\
    F1  &SQuAD &$D_{\tt TV}$  &\xmark          &0.490 &0.393 &0.286 &0.261 &0.475 &0.863 \\
    F1  &SQuAD &$D_{\tt HL}$  &\xmark          &0.481 &0.387 &0.274 &0.255 &0.487 &0.862 \\
    \hdashline
    Sum $y$&SQuAD &$D_{\tt KL}$&$\mathcal{N}_y$&0.483 &0.396 &0.229 &0.249 &0.545 &0.943 \\
    Sum $y$&SQuAD &$D_{\tt OB}$&$\mathcal{N}_y$&0.517 &0.385 &0.286 &0.256 &0.711 &0.914 \\
    Sum $y$&SQuAD &$D_{\tt TV}$&$\mathcal{N}_y$&0.519 &0.407 &0.324 &0.292 &0.502 &0.890 \\
    Sum $y$&SQuAD &$D_{\tt HL}$&$\mathcal{N}_y$&0.512 &0.413 &0.323 &0.299 &0.385 &0.889 \\
    \hdashline
    Src $x$&RACE &$D_{\tt KL}$ &\xmark         &0.143 &0.097 &0.088 &0.054 &0.321 &0.599  \\
    Src $x$&RACE &$D_{\tt OB}$ &\xmark         &0.226 &0.091 &0.160 &0.091 &0.534 &0.612 \\
    Src $x$&RACE &$D_{\tt TV}$ &\xmark         &0.233 &0.143 &0.069 &0.087 &0.144 &0.588 \\
    Src $x$&RACE &$D_{\tt HL}$ &\xmark         &0.221 &0.148 &0.056 &0.083 &0.222 &0.592 \\
    \hdashline
    Sum $y$&RACE &$D_{\tt KL}$ &\xmark         &0.450 &0.283 &0.135 &0.179 &0.789 &0.954\\
    Sum $y$&RACE &$D_{\tt OB}$ &\xmark         &0.453 &0.225 &0.240 &0.221 &0.839 &0.928\\
    Sum $y$&RACE &$D_{\tt TV}$ &\xmark         &0.462 &0.309 &0.221 &0.244 &0.770 &0.933\\
    Sum $y$&RACE &$D_{\tt HL}$ &\xmark         &0.473 &0.323 &0.215 &0.244 &0.751 &0.927\\
    \hdashline
    F1  &RACE &$D_{\tt KL}$ &\xmark            &0.480 &0.266 &0.156 &0.198 &0.830 &0.908 \\
    F1  &RACE &$D_{\tt OB}$ &\xmark            &0.379 &0.192 &0.268 &0.206 &0.796 &0.815 \\
    F1  &RACE &$D_{\tt TV}$ &\xmark            &0.468 &0.301 &0.217 &0.252 &0.731 &0.866 \\
    F1  &RACE &$D_{\tt HL}$ &\xmark            &0.472 &0.317 &0.206 &0.252 &0.693 &0.858 \\
    \hdashline
    Sum $y$&RACE &$D_{\tt KL}$&$\mathcal{N}_y$ &0.460 &0.302 &0.208 &0.206 &0.857 &0.961 \\
    Sum $y$&RACE &$D_{\tt OB}$&$\mathcal{N}_y$ &0.466 &0.233 &0.266 &0.226 &0.822 &0.954 \\
    Sum $y$&RACE &$D_{\tt TV}$&$\mathcal{N}_y$ &0.502 &0.313 &0.306 &0.270 &0.855 &0.945 \\
    Sum $y$&RACE &$D_{\tt HL}$&$\mathcal{N}_y$ &0.501 &0.328 &0.305 &0.273 &0.860 &0.936 \\
    \bottomrule
  \end{tabular}
  \caption{Pearson correlation coefficients of all MQAG configurations. Our MQAG results are based on $N$=50. When applying the answerability mechanism, the threshold $\mathcal{N}^{\tau}_y$ is set to 2.0.}
  \label{tab:mqag_pearson_appendix}
\end{table*}

\begin{table*}[!t]
  \centering
\scalebox{0.96}{
  \begin{tabular}{lcccccc}
    \toprule
    \multirow{2}{*}{Method} &\multicolumn{2}{c}{QAG} &\multicolumn{2}{c}{XSum-H} &\multirow{2}{*}{Podcast} &\multirow{2}{*}{SumEvl} \\
     &CNNDM &XSum &Faithful &Factual  &  \\
    \midrule
    \rowcolor{Gray}
    \multicolumn{7}{l}{\textit{Baselines: Other Approaches}} \\
    ROUGE-1                  &0.318 &0.053	&-0.030	&0.001	&0.282	&0.627  \\
    OpenIE-TripleMatching    &0.337	&0.130	&0.019	&-0.025	&0.700	&0.671  \\
    BERTScore                &0.523	&0.018	&0.183	&0.153	&0.686	&0.835  \\
    Entailment (BERT Model)  &0.167	&0.190	&0.380	&0.202	&0.207	&0.141  \\
    \rowcolor{Gray}
    \multicolumn{7}{l}{\textit{Baselines: SpanQAG}} \\
    QAGS                     &0.341	 &0.166	&0.085	&0.052	&0.357	&0.421  \\
    FEQA                     &0.275	 &0.277	&0.300	&0.155	&0.504	&0.270  \\
    QuestEval                &0.181	 &0.175	&0.415	&0.176	&0.425	&0.812  \\
    \rowcolor{Gray}
    \multicolumn{7}{l}{\textbf{Multiple-choice Question Answering and Generation (MQAG)}} \\
    MQAG$_\text{SQuAD}$      &0.470	 &0.409	&0.335	&0.284	&0.441	&0.773 \\ 
    MQAG$_\text{RACE}$       &0.460	 &0.308	&0.322	&0.266	&0.779	&0.920 \\
    \bottomrule
  \end{tabular}}
  \caption{Spearman's rank correlation coefficient between the scores of summary evaluation methods and human judgements. This table is complementary to Table~\ref{tab:baseline_vs_mqag} which reports Pearson's correlation coefficient results.}
  \label{tab:baseline_vs_mqag_spearman}
\end{table*}

\begin{table*}[ht]
  \centering
  \scalebox{0.92}{
  \begin{tabular}{p{\linewidth}}
    \toprule
    \textbf{Source}: A G4S security van has been robbed outside a branch of royal bank of Scotland in \underline{Glasgow city centre}. Police said three armed men took a five-figure sum from the vehicle in the city's Sauchiehall street on Monday at about 21:45. A spokesman said no-one had been injured although two security guards aged 47 and 49 were left badly shaken. The area around the bank, which is near the Buchanan galleries shopping centre, has been cordoned off by police. Police said the security guards had been making their delivery when they were approached by the three armed men, who threatened them and demanded they hand over a box of money. It is understood the cash taken was in the region of £50,000. Following the robbery, the three men got into a white seat Leon car, which sped off along west Nile street towards the cowcaddens area. [...] \\
    
    \midrule
    \textbf{Summary}: Two security guards have been threatened during a robbery at a bank in \textcolor{red}{Edinburgh}. \\
    \midrule
    \textbf{Generated question (using summary)}: The robbery happened in \_ . \\
    \textbf{Generated options (using summary)}: (1) Edinburgh (2) a bank (3) a shop (4) a small town. \\
    \midrule
    \textbf{Prob. over options given Source}: \texttt{ 0.077, 0.895, 0.018, 0.010} \\ 
    \textbf{Prob. over options given Summary}: \texttt{0.687, 0.295, 0.000, 0.018} \\ 
    \bottomrule
  \end{tabular}}
  \caption{Example from QAG-XSum (documentID=1). Factual inconsistency in the summary is highlighted in red.}
  \label{tab:example1}
\end{table*}

\end{document}